\documentclass[10pt,twocolumn,letterpaper]{article}

\usepackage{cvpr}              %

\usepackage{graphicx}
\usepackage{amsmath}
\usepackage{amssymb}
\usepackage{booktabs}
\usepackage{tabularx}
\usepackage{array} 
\usepackage{multirow}
\usepackage{cite}
\usepackage[export]{adjustbox}
\usepackage{xcolor}
\usepackage{makecell}
\usepackage[accsupp]{axessibility}

\definecolor{cvprblue}{rgb}{0.21,0.49,0.74}
\usepackage[pagebackref,breaklinks,colorlinks,citecolor=cvprblue]{hyperref}

\usepackage[capitalize]{cleveref}
\crefname{section}{Sec.}{Secs.}
\Crefname{section}{Section}{Sections}
\Crefname{table}{Table}{Tables}
\crefname{table}{Tab.}{Tabs.}

\newcommand{\bx}{\mathbf{x}}
\newcommand{\by}{\mathbf{y}}

\newcommand{\bz}{\mathbf{z}}

\newcommand{\bc}{\mathbf{c}}

\newcommand{\bd}{\mathbf{d}}

\newcommand{\nR}{\mathbb{R}}

\newcommand{\nS}{\mathbb{S}}

\makeatletter
\DeclareRobustCommand\onedot{\futurelet\@let@token\@onedot}
\def\@onedot{\ifx\@let@token.\else.\null\fi\xspace}
\def\eg{e.g\onedot}

\def\etc{etc\onedot}

\def\Fig{Fig\onedot}   
\makeatother

\newcommand{\figref}[1]{\Fig~\ref{#1}}
\newcommand{\secref}[1]{Section~\ref{#1}}

\renewcommand{\eqref}[1]{Eq.~\ref{#1}}
\newcommand{\tabref}[1]{Table~\ref{#1}}

\newcommand{\boldparagraph}[1]{\vspace{0.2cm}\noindent{\bf #1:} }
\newcommand{\qparagraph}[1]{\vspace{0.2cm}\noindent{\bf #1} }

\newif\ifcomment
\commenttrue
\ifcomment
	\newcommand{\ag}[1]{ \noindent {\color{red} {\bf Andreas:} {#1}} }
	\newcommand{\yl}[1]{ \noindent {\color{cyan} {\bf YL:} {#1}} }

\else
	\newcommand{\ag}[1]{}
	\newcommand{\yl}[1]{}
\fi

\begin{document}

\title{NeRFCodec: Neural Feature Compression Meets Neural Radiance Fields for Memory-Efficient Scene Representation}

\makeatletter
\def\blfootnote{\gdef\@thefnmark{}\@footnotetext}
\makeatother

\author{Sicheng Li
\qquad Hao Li 
\qquad Yiyi Liao$^{*}$
\qquad Lu Yu$^{**}$
\vspace{1em}
\\
Zhejiang University
}
\maketitle

\begin{abstract}
The emergence of Neural Radiance Fields (NeRF) has greatly impacted 3D scene modeling and novel-view synthesis. As a kind of visual media for 3D scene representation, compression with high rate-distortion performance is an eternal target. Motivated by advances in neural compression and neural field representation, we propose NeRFCodec, an end-to-end NeRF compression framework that integrates non-linear transform, quantization, and entropy coding for memory-efficient scene representation. Since training a non-linear transform directly on a large scale of NeRF feature planes is impractical, we discover that pre-trained neural 2D image codec can be utilized for compressing the features when adding content-specific parameters. Specifically, we reuse neural 2D image codec but modify its encoder and decoder heads, while keeping the other parts of the pre-trained decoder frozen. This allows us to train the full pipeline via supervision of rendering loss and entropy loss, yielding the rate-distortion balance by updating the content-specific parameters. At test time, the bitstreams containing latent code, feature decoder head, and other side information are transmitted for communication. Experimental results demonstrate our method outperforms existing NeRF compression methods, enabling high-quality novel view synthesis with a memory budget of 0.5 MB.
\end{abstract}

\blfootnote{$^*$ Corresponding author. $^{**}$ Co-corresponding author.}
\section{Introduction}
Neural Radiance Fields~(NeRF)~\cite{mildenhall2020nerf} have emerged as a popular scene representation for novel view synthesis.
As a promising representation for immersive media, how to compress NeRF with a better storage-quality trade-off is a significant problem for efficient communication and storage.

While the deep MLPs used in NeRF~\cite{mildenhall2020nerf} are parameter-efficient, the hybrid representation~\cite{DVGO, yu2021plenoxels}, which combines feature grids and small MLPs, has become the mainstream due to its high reconstruction quality, fast training speed, and efficient rendering. 
However, its drawback lies in the significant storage requirements. 
Consequently, subsequent research efforts have emerged to reduce the storage footprint of the hybrid representation without compromising reconstruction quality.

\begin{figure}
    \centering
    \begin{tabular}{cc}
    \resizebox{4.5cm}{!}{\includegraphics[width=\linewidth]{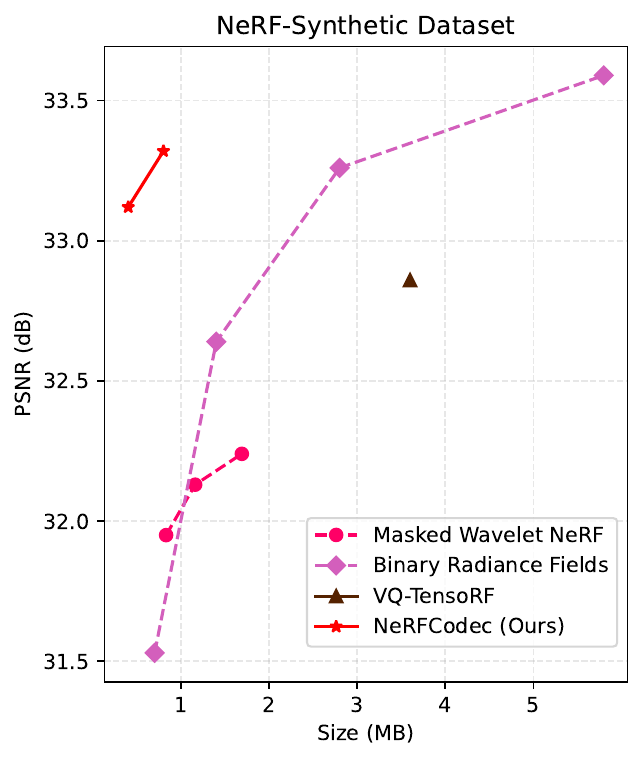}} & 
    \resizebox{3cm}{!}{\includegraphics[width=\linewidth]{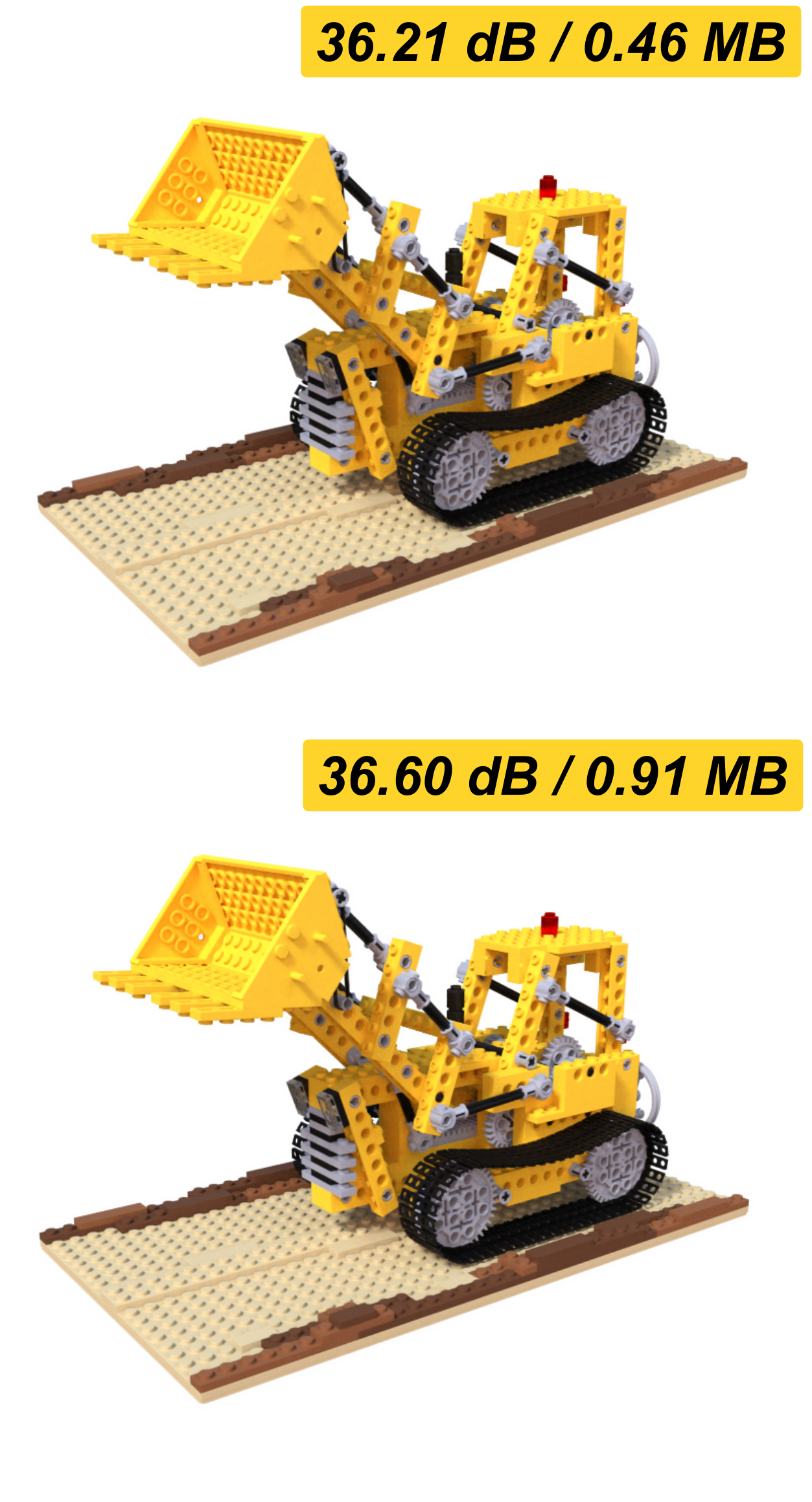}}
    \\ 
    (a)~Rate-distortion performance & (b)~Rendered images
    \end{tabular}
    \vspace{-0.3cm}
    \caption{\textbf{Compression performance}.}
    \label{fig:teaser}
    \vspace{-0.7cm}
\end{figure}

One line of work focuses on efficient data structure design of feature grids, which involves using more parameter-efficient data structures, e.g.,  tensor factorization-based representations~\cite{Chen2022ECCV, eg3dChanLCNPMGGTKKW22} and multi-resolution hash grids~\cite{muller2022instant}, to replace dense voxel grids.
These methods allow for reducing the number of parameters, e.g., from 1GB to 50MB, without sacrificing the quality. However, this size still requires further reduction.

Another line of work focuses on compressing parameters using compression techniques like quantization~\cite{takikawa2022vqad,li2023vqrf, shin2023birf} and entropy coding~\cite{li2023vqrf}.
However, most works in this field overlook another effective compression method, transform coding~\cite{goyal2001transformcoding}, that transforms the original data to another space based on linear~\cite{lewis1992image, watson1994image} or non-linear mapping~\cite{balle2020nonlinear}. The image compression community demonstrates that the combination of transform coding, quantization, and entropy coding leads to the most competitive compression performance~\cite{skodras2001jpeg, sullivan2012hevcoverview, bross2021vvcoverview}.
Recently, there is a work named Masked Wavelet NeRF~\cite{rho2023maskedwavelet}, which is the first to consider transform coding together with quantization and entropy coding in NeRF compression and achieves promising performance. 
Despite the success, Masked Wavelet NeRF only leverages a linear transform, which has proved to be less effective than learned non-linear transform in neural image compression~\cite{cheng2020learned, he2022elic, hu2021learning}.

Therefore, motivated by advances in neural image compression, we introduce NeRFCodec, a NeRF compression framework that integrates non-linear transform, quantization, and entropy coding for compressing feature planes in hybrid NeRF to achieve memory-efficient scene representation. A key question is, how do we obtain the non-linear basis? Existing 2D neural image compression methods obtain such a non-linear transform by training an encoder-decoder network on millions of images, while it is infeasible to train such a neural codec on millions of 3D scenes. We find that, surprisingly, existing neural image codec trained on natural images can serve as a strong backbone for compressing feature planes of NeRF, when partially tailored and tuned to each scene. Specifically, we first pre-train a hybrid NeRF on one scene, and feed the feature planes into a well-trained 2D image neural codec by replacing its encoder and decoder heads (i.e., the first and last layers) to adjust to the target channel dimension. Next, we adapt the full encoder and the decoder head to fit this scene while keeping the remaining parts of the decoder frozen.  This is supervised by rate-distortion loss, with the goal of decoding the feature planes for high image rendering quality while maintaining a low bitrate of the latent code via an entropy loss. After training, the latent code predicted by the encoder and the decoder head parameters are quantized and entropy-coded into a bitstream for transmission. 

Our experimental results demonstrate that NeRFCodec pushes the frontier of the rate-distortion trade-off compared to existing NeRF compression methods~\cite{li2023vqrf, rho2023maskedwavelet, shin2023birf}. Our method only uses 0.5 MB to represent a single scene while maintaining high visual fidelity, as shown in \figref{fig:teaser}.

We summarize our contributions as follows: 

\begin{itemize}
    \item We introduce NeRFCodec, an end-to-end compression framework for plane-based hybrid NeRF. It utilizes neural feature compression, combining non-linear transformation, quantization, and entropy coding for efficient compression of plane-based NeRF representations, which advances the frontier of rate-distortion performance for compact NeRF representations.
    \item We propose to re-use pre-trained neural 2D image codec with slight modification and fine-tune it to each scene individually via the supervision of rate-distortion loss. 
    \item We demonstrate that our method could achieve superior rate-distortion performance compared to existing NeRF compression methods. 
\end{itemize}

\section{Related Work}
\boldparagraph{Efficient Representation of Neural Fields}
The vanilla NeRF~\cite{mildenhall2020nerf} proposes to represent the scene with a multi-layer perception~(MLP). Subsequent works demonstrate that representing the scene as voxel grids leads to significantly faster training and better reconstruction quality~\cite{yu2021plenoxels, DVGO, karnewar2022relufields}.
Besides, point-based methods~\cite{xu2022pointnerf, zhang2022differentiable}, like Point-NeRF~\cite{xu2022pointnerf}, demonstrated its capability for high-efficiency NeRF reconstruction.
However, 3D dense voxel grids and point clouds require substantial memory.

Several strategies~\cite{muller2022instant, tang2022ccnerf, eg3dChanLCNPMGGTKKW22, Chen2022ECCV, fridovich2023kplanes, deng2023renerf} have been proposed for efficient scene representation design to alleviate the stress of memory requirements.
Instant-NGP~\cite{muller2022instant} reduces the memory cost of high-resolution voxel grids by constructing a hash encoding and resolving hash collision implicitly by a tiny MLP decoder.
Another line of work decomposes 3D feature volumes into orthogonal 2D planes or 1D vectors. 
EG3D~\cite{eg3dChanLCNPMGGTKKW22} introduces a tri-plane representation of three perpendicular feature planes and extracts features separately from each plane as inputs for the following MLPs.
TensoRF~\cite{Chen2022ECCV} is inspired by tensor decomposition to represent 3D grids with combinations of axis-aligned vectors and matrices via VM decomposition and CP decomposition.
While these strategies successfully reduce the parameter count without compromising rendering quality, their raw uncompressed parameters still require at least tens of megabytes for storage.

\begin{figure*}
  \centering
   \includegraphics[width=\linewidth]{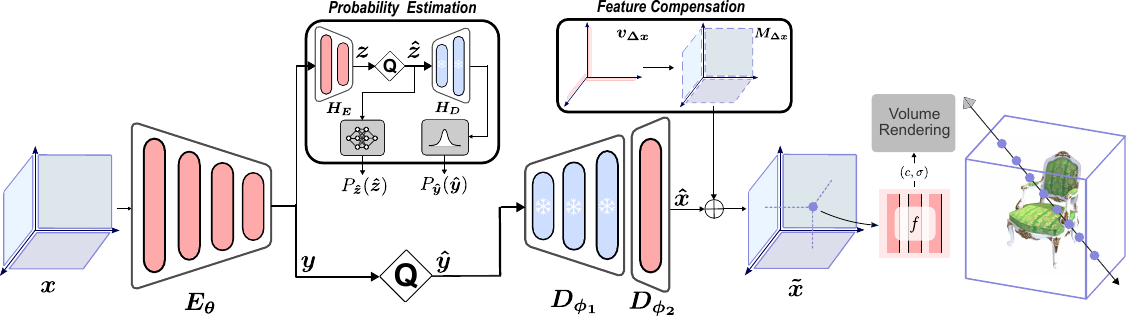}
   \caption{\textbf{NeRFCodec}. We combine the pre-trained neural 2D image codec with content-specific parameters to compress hybrid NeRF. The feature planes $\boldsymbol{x}$ are fed into feature encoder $\boldsymbol{E_{\theta}}$ to obtain latent code $\boldsymbol{y}$. Latent code $\boldsymbol{y}$ is quantized into $\boldsymbol{\hat{y}}$ in one branch. In another branch, latent code $\boldsymbol{y}$ is sent to the probability estimation module to get its corresponding probability $P_{\boldsymbol{\hat{y}}}$ of quantized latent code $\boldsymbol{\hat{y}}$ for entropy coding. Inside the probability estimation, it leverages a hyperprior encoder $\boldsymbol{H_{E}}$ to obtain hyperprior latent code $\boldsymbol{{z}}$ and a hyperprior decoder $\boldsymbol{H_{D}}$ to estimate probability distribution $P_{\boldsymbol{\hat{y}}}$. The quantized latent code $\boldsymbol{\hat{y}}$ is fed into the feature decoder $\boldsymbol{D_{\phi}}$ to generate reconstructed feature planes $\boldsymbol{\hat{x}}$. The feature decoder consists of feature decoder backbone $\boldsymbol{D_{\phi_{1}}}$ and feature decoder head $\boldsymbol{D_{\phi_{2}}}$. We introduce a feature compensation module to compensate for the loss of high-frequency residuals. We add feature residual matrix $\boldsymbol{M_{\Delta \boldsymbol{x}}}$ represented by the outer product of feature residual vectors $\boldsymbol{v_{\Delta \boldsymbol{x}}}$ to get the final feature planes $\boldsymbol{\tilde{x}}$. The final feature planes $\boldsymbol{\tilde{x}}$ cooperate with a tiny MLP $f$ to predict the color and density of sample points for volume rendering. The red components are updated in training, while the blue components inherit parameters from pre-trained neural image compression and stay frozen. The final bitstreams include quantized latent code $\boldsymbol{\hat{y}}$, quantized hyperprior latent code $\boldsymbol{\hat{z}}$, feature decoder head $\boldsymbol{D_{\phi_{2}}}$, feature residual vectors $\boldsymbol{v_{\Delta \boldsymbol{x}}}$, tiny MLPs $f$, and metadata.}
   \label{fig:pipeline}
   \vspace{-0.5cm}
\end{figure*}

\boldparagraph{Compression of Neural Fields Representation}
Various compression techniques have been applied to the aforementioned representations of neural fields for further compression, including parameter quantization techniques, transform coding, and entropy coding.

Quantization techniques could be categorized into vector quantization and scalar quantization in terms of the parameter unit to be quantized. 
VQAD~\cite{takikawa2022vqad} introduces vector quantization to compress tree-based neural field representations~\cite{takikawa2021nglod}.
This approach adaptively learns features within the codebook and the index assignment corresponding to various leaf nodes during training.
VQRF~\cite{li2023vqrf} introduces a universal compression pipeline designed for pre-trained hybrid NeRF representations, including voxel pruning, vector quantization, weight quantization, and entropy coding.
Masked Wavelet NeRF~\cite{rho2023maskedwavelet} employs quantization-aware training and 8-bit uniform scalar quantization on wavelet coefficients.
BiRF~\cite{shin2023birf} introduces the concept of binary neural networks~\cite{courbariaux2015binaryconnect} into the NeRF domain. It uses hash-encoded 2D planes and 3D volumes as scene representations, where each feature is quantized to either +1 or -1.

Entropy coding is a way to achieve lossless compression of a sequence of symbols, which is the foundation of advanced compression systems.
cNeRF~\cite{bird2021cnerf} individually learns a probability model for employing entropy coding to weights of MLP in the vanilla NeRF, which follows the strategy of prior neural network compression work~\cite{oktay2019scalable}. 
VQRF utilizes entropy coding in the final step to compress simplified components of the entire model individually and then pack the bitstream together. 
Masked Wavelet NeRF performs run-length encoding (RLE) to masked wavelet coefficients and applies the Huffman encoding to the RLE-encoded streams to map values with a high probability to shorter bits.

Transform coding is an effective lossy compression technique demonstrated in traditional image and video coding.
Masked Wavelet NeRF takes inspiration from JPEG~2000~\cite{skodras2001jpeg}, employing inverse wavelet transform on learnable 1D vectors and 2D planes to obtain features for subsequent queries. ACRF~\cite{fang2024acrf} is motivated by point cloud compression and introduces a point-based wavelet transform, region adaptive hierarchical transform (RAHT)~\cite{de2016RAHT}, for voxel/point-based NeRF compression.

Our approach follows the compression framework that combines transform coding, quantization, and entropy coding. Compared to previous works, we introduce a non-linear transform designed for the feature planes in hybrid NeRF.

\boldparagraph{Neural Image Compression}
Recent research has made rapid progress on deep learning-based neural image compression methods. The current state-of-the-art methods can approach or even surpass advanced traditional image and video codecs in terms of rate-distortion performance.
Image coding standards based on the neural image compression framework, JPEG-AI~\cite{ascenso2023jpegai}, are also under development.
The main paradigm for neural image compression is the autoencoder, which inserts a scalar quantization and an entropy coding module into the bottleneck layer.
The auto-encoder is regarded as a pair of non-linear transform basis, while the latent code is regarded as non-linear transform coefficients~\cite{balle2020nonlinear}.
An essential work proposed by Balle et al.~\cite{balle2018hyperprior} introduces a hyperprior network for better probability estimation of latent code in entropy coding with minor overhead of sending side information, which becomes an indispensable building block in neural image compression.
Subsequently, several methods were proposed to enhance the rate-distortion performance of neural image compression, including more expressive networks in non-linear transform~\cite{cheng2020learned, liu2023learned}, more precise probability estimation~\cite{cheng2020learned, minnen2018joint}, and generative adversarial training~\cite{mentzer2020hific}.

Despite superior natural image compression performance, directly applying these advanced methods to compressing feature planes in hybrid NeRF without modification does not achieve high rate-distortion performance, as demonstrated in our experimental section.

\section{Method}

In this work, we propose an end-to-end NeRF compression framework 
compatible with plane-based hybrid NeRF variants. 
\figref{fig:pipeline} gives an overview of our framework, comprising neural feature compression and NeRF rendering. 
Neural feature compression consists of content-adaptive non-linear transform, quantization, and entropy coding. 
NeRF rendering follows the corresponding NeRF variants.

In the following, we first introduce preliminaries of hybrid NeRF model and neural image compression in \secref{sec:nerf}. Then, we provide a preliminary toy experiment in \secref{sec:analysis} to assess the feasibility of our methods. Next, we illustrate neural feature compression for NeRF representation compression in \secref{sec:codec}, and the training strategy in \secref{sec:training}. Furthermore, we describe the encoding and decoding process at the test time in \secref{sec:test-time}. Finally, we describe implementation details in \secref{sec:detail}. 

\subsection{Background}
\label{sec:nerf}
\boldparagraph{NeRF}
Generally, NeRF could be regarded as a continuous scene representation function $g_\theta$ parameterized by learnable parameters $\theta$ that maps the location of a 3D point $\bx\in \nR^3$ and a viewing direction $\bd \in \nS^2$ towards the point to a volume density $\sigma$ and a color value $\bc$:
\begin{equation}
    g_{\theta}: (\bx\in\nR^3, \bd\in\nS^2) \mapsto (\sigma\in\nR^+,\bc\in\nR^3)
\end{equation}

Given a target camera, the color $\bc_r$ of a target pixel corresponding to a camera ray $r$ is obtained via volume rendering integral approximated by the numerical quadrature:
\begin{gather}
    \bc_r =\sum_{i=1}^{N} T_r^i \alpha_r^i \bc_r^i \   \\
    \alpha_r^i = 1-\exp (-\sigma_r^i\delta_r^i) \quad  T_r^i   = \prod_{j=1}^{i-1}\left(1-\alpha_r^j\right)
\end{gather}
where $T_r^i$ and $\alpha_r^i$ denote transmittance and alpha value of a sample point $\bx_i$ along the camera ray $r$. 

Hybrid NeRF consists of explicit data structures and tiny MLP $f$. In the rendering process of hybrid NeRF, we first query features from explicit data structure based on the location of sample points, and then we use the tiny MLP $f$ to map queried features to final density and color. The explicit data structure accounts for over 95$\%$ of the total parameters and often amounts to tens or even hundreds of megabytes, regarded as the main target for compression.
In this paper, we focus on leveraging neural compression to compress plane-based hybrid NeRF, a parameter-efficient and widely used variant of hybrid NeRF.

\boldparagraph{Neural Image Compression}
The basic framework of neural image compression~\cite{balle2018hyperprior, he2022elic, cheng2020learned} consists of non-linear transforms parameterized by a pair of encoder~$\boldsymbol{E}$ and decoder~$\boldsymbol{D}$, quantization~$\boldsymbol{Q}$, and entropy coding with a learned probability estimation module.

The image $\boldsymbol{x}$ is fed into the encoder~$\boldsymbol{E}$ to obtain low-dimensional latent code $\boldsymbol{y}$, which would be quantized into $\boldsymbol{\hat{y}}$.
Then, the quantized latent code $\boldsymbol{\hat{y}}$ is fed into the decoder~$\boldsymbol{D}$ to obtain reconstructed image $\boldsymbol{\hat{x}}$.
\begin{equation}
    \boldsymbol{\hat{y}} = \boldsymbol{Q}(\boldsymbol{E}(\boldsymbol{x})) \quad \boldsymbol{\hat{x}} = \boldsymbol{D}(\boldsymbol{\hat{y}})
\end{equation}
For better modeling the probability of $\boldsymbol{\hat{y}}$, the hyperprior latent code $\boldsymbol{z}$ is introduced as side information via a hyperprior auto-encoder.
$\boldsymbol{z}$ is obtained by feeding $\boldsymbol{y}$ into a hyperprior encoder $\boldsymbol{H_{E}}$.Then, $\boldsymbol{z}$ is quantized and fed into a hyperprior decoder $\boldsymbol{H_{D}}$ to recover parameters for probability distribution modeling of $P_{\boldsymbol{\hat{y}}}$. The quantized hyperprior latent code $\boldsymbol{\hat{z}}$ must also be entropy-coded and transmitted. 
\begin{equation}
    \boldsymbol{\hat{z}} = \boldsymbol{Q}(\boldsymbol{H_{E}}(\boldsymbol{y})) \quad P_{\boldsymbol{\hat{y}}} \leftarrow \boldsymbol{H_{D}}(\boldsymbol{\hat{z}})
\end{equation}

The probability distribution $P_{\boldsymbol{\hat{y}}}$ of quantized latent code $\boldsymbol{\hat{y}}$ is a Gaussian distribution conditioned on $\boldsymbol{\hat{z}}$. There is no prior for $P_{\hat{\bz}}$, so the probability distribution $P_{\boldsymbol{\hat{z}}}$ of quantized hyperprior latent code is modeled by a factorized density model parameterized by shallow MLP.

In neural image compression, once the neural image codec is trained on millions of natural images, the architecture and parameters of the decoder are fixed. The pre-trained decoder is assumed to be capable of decoding any entropy-coded bitstream that meets standard requirements. As a result, the cost of transmitting the decoder is amortized over countless bitstreams. Therefore, when calculating the amount of data to be transmitted, only the size of the entropy-coded latent code needs to be considered.

\subsection{Preliminary Analysis}
\label{sec:analysis}
Although large-scale datasets~\cite{deitke2023objaverse, yu2023mvimgnet, wu2023omniobject3d} containing millions of 3D scenes have been released, it is infeasible to train feature planes with millions of 3D data and then train a feature codec with millions of feature planes, considering time and resource consumption.
Alternatively, we investigate the possibility of reusing a neural 2D image codec trained on millions of natural images and applying it to compress feature planes in hybrid NeRF with slight modification.

\begin{figure}[tb]
  \centering
   \includegraphics[width=\linewidth]{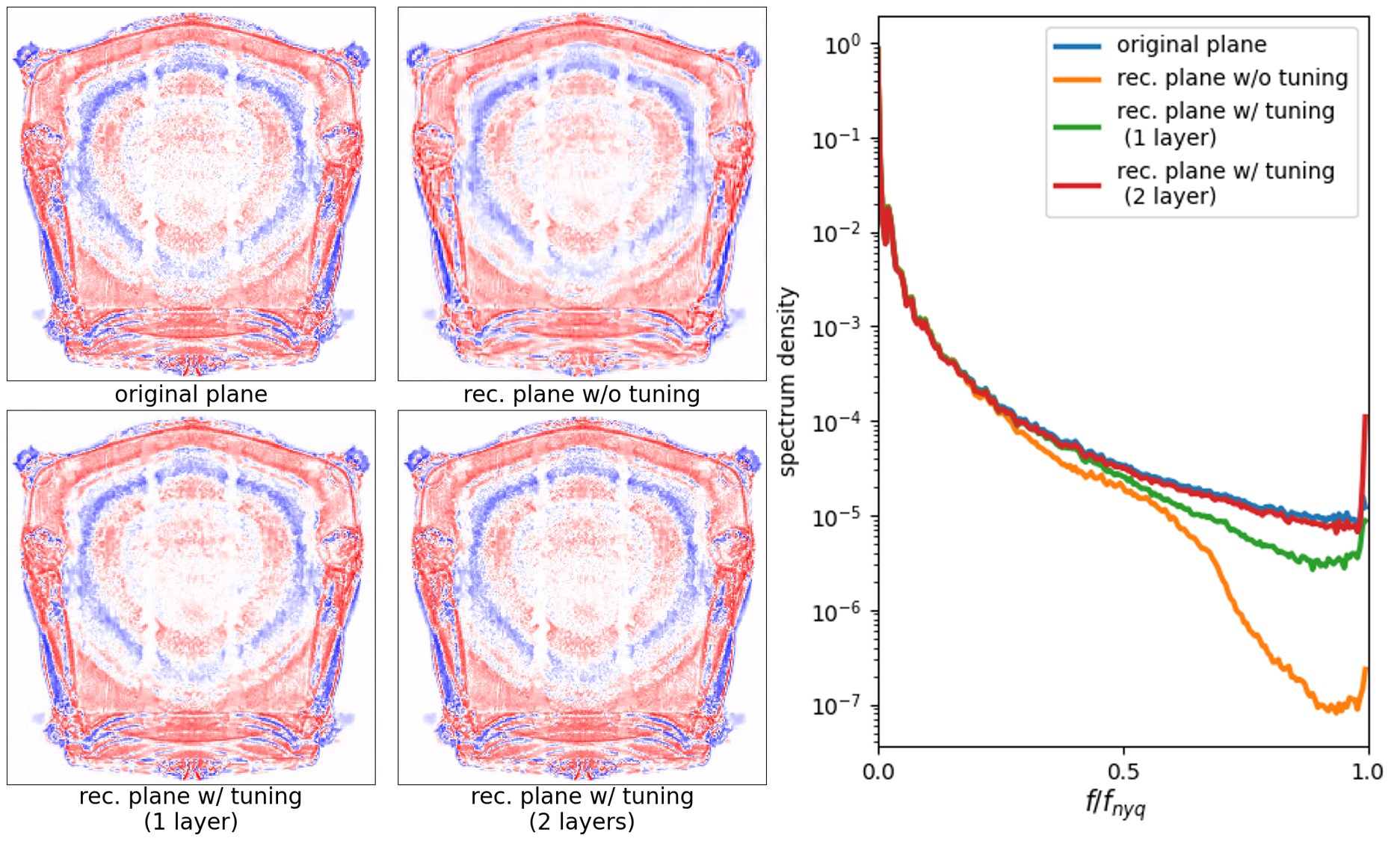}
   \vspace{-0.7cm}
   \caption{\textbf{Spectrum analysis of decoded feature plane}.}
   \label{fig:spectrum}
   \vspace{-0.4cm}
\end{figure}

Thus, we design a toy experiment to assess the feasibility of our approach in a simple manner.
We extract three channels of pre-trained feature planes of TensoRF~\cite{Chen2022ECCV}, normalize their value range to be between 0 and 1, and feed them into a pre-trained neural 2D image codec.
We visualize the reconstructed feature map and analyze its spectrum in \figref{fig:spectrum}. Unsurprisingly, the na\"ive way of resuing the 2D image codec suffers from domain gap. This leads to a significant energy loss in the high-frequency range compared to the original.
Next, we fine-tune the last layer of the decoder to reconstruct the original feature plane, while keeping the other layers frozen. We find that simply fine-tuning the last layer recovers the high-frequency energy in the reconstructed feature map to a large extent. This allows for a receiver to decode the feature plane from the latent code, when the content-adapted last layer is jointly transmitted.  While fine-tuning more layers in the decoder further improves the performance, it can lead to a significant increase in the data transmission cost.

\subsection{NeRFCodec}
\label{sec:codec}
Based on previous analysis, we designed our pipeline, as shown in \figref{fig:pipeline}.
The neural feature compression in our pipeline involves a content-adaptive feature encoder $\boldsymbol{E_{\theta}}$, a feature decoder~$\boldsymbol{D_{\phi}}$ with a tuned 
content-adaptive decoder head~$\boldsymbol{D_{\phi_{2}}}$, a feature compensation module and other indispensable components in neural compression including quantization and entropy coding. 

\boldparagraph{Content-Adaptive Encoder}
Our feature encoder $\boldsymbol{E_{\theta}}$ inherits the pre-trained encoder in the neural 2D image codec, except that it replaces the first layer of the pre-trained model with a new encoder head to adapt to the channel dimension of feature planes.
In brief, we initially encode each feature plane $\boldsymbol{x}$ individually through content-adaptive encoder $\boldsymbol{E_{\theta}}$ into a latent code $\boldsymbol{y}$.
\begin{equation}
    \boldsymbol{y} = \boldsymbol{E_{\theta}}(\boldsymbol{x})
\end{equation}

During training, all parameters $\boldsymbol{\theta}$ of our feature encoder are optimized, leading to a content-adaptive feature encoder for each scene individually.
Since our encoder is customized for scene-specific optimization, the resulting latent code obtained in this way is also optimized for each scene. While it is possible to optimize the latent code directly without the need for an encoder, our ablation study indicates that the encoder-free approach is less competitive.

\boldparagraph{Content-Adaptive Decoder Head}
Our feature decoder~$\boldsymbol{D_{\phi}}$ comprises feature decoder backbone ~$\boldsymbol{D_{\phi_{1}}}$ and feature decoder head~$\boldsymbol{D_{\phi_{2}}}$.
Similar to the neural feature encoder, we also re-use the decoder in pre-trained neural image codec except for the last layer and initialize a new final layer to adjust to the number of target channel dimensions. The feature decoder takes quantized latent code~$\boldsymbol{\hat{y}}$ as input and outputs reconstructed feature planes~$\boldsymbol{\hat{x}}$.
\begin{equation}
    \boldsymbol{\hat{x}} = \boldsymbol{D_{\phi}}(\boldsymbol{\hat{y}}) = \boldsymbol{D_{\phi_{2}}}(\boldsymbol{D_{\phi_{1}}}(\boldsymbol{\hat{y}}))
\end{equation}

In training, the feature decoder backbone~$\boldsymbol{D_{\phi_{1}}}$ stay fixed while the parameters in feature decoder head~$\boldsymbol{D_{\phi_{2}}}$ are updated. 
Thus, the parameters in feature decoder head~$\boldsymbol{D_{\phi_{2}}}$ need to be compressed and transmitted to the receiver side for decoding the feature planes correctly.
The feature decoder head can be optimized jointly with the MLP $f$ in hybrid NeRF, effectively predicting the attributes of each point in the scene with high quality.

\boldparagraph{Feature Compensation}
Lossy compression leads to the loss of high-frequency details, as shown in \secref{sec:analysis}.
Thus, we introduce a high-frequency residual compensation module $\boldsymbol{C_{R}}$ tailored for decoded feature planes $\boldsymbol{\hat{x}}$. 
This method aims to increase high-frequency details on the reconstructed feature planes with a low storage cost, thereby enhancing the final rendering quality.
In practice, we assign a residual matrix~$\boldsymbol{M_{\Delta \boldsymbol{x}}}$ to each reconstructed feature plane $\boldsymbol{\hat{x}}$ for compensating high-frequency details. To avoid the burden of directly storing the matrix, we further represent this high-frequency compensation matrix as the outer product of two orthogonal factorized vectors~$\boldsymbol{v_{\Delta \boldsymbol{x}}}$ with learnable parameters, following CP decomposition in TensoRF. 
The final feature plane~$\boldsymbol{\tilde{x}}$ used for feature queries is the sum of the reconstructed feature plane $\boldsymbol{\hat{x}}$ and the residual plane~$\boldsymbol{M_{\Delta \boldsymbol{x}}}$ parameterized by factorized vectors~$\boldsymbol{v_{\Delta \boldsymbol{x}}}$.
\begin{equation}
\begin{aligned}
    \boldsymbol{\tilde{x}} = \boldsymbol{\hat{x}} + \boldsymbol{M_{\Delta \boldsymbol{x}}} 
                           = \boldsymbol{\hat{x}} + \boldsymbol{v^i_{\Delta \boldsymbol{x}}} \circ \boldsymbol{v^j_{\Delta \boldsymbol{x}}}
\end{aligned}
\end{equation}
At test time, these feature vectors~$\boldsymbol{v_{\Delta \boldsymbol{x}}}$ will be quantized and entropy-coded into the bitstreams for transmission.

\boldparagraph{Quantization}
A scalar quantizer $\boldsymbol{Q}$ is introduced to quantize latent code $\boldsymbol{y}$ into $\boldsymbol{\hat{y}}$.
While quantization is not differentiable, we follow the protocol strategy in neural image compression~\cite{cheng2020learned, balle2018hyperprior} to add uniform noise to the latent code to simulate quantization in training. 

\boldparagraph{Entropy Coding}
We follow the protocols of entropy coding in neural image compression~\cite{cheng2020learned, balle2018hyperprior}.
An essential operation in entropy coding is symbol probability estimation.
In training, the estimated probability $P_{\boldsymbol{\hat{y}}}$ and $P_{\boldsymbol{\hat{z}}}$ is used to calculate their entropy~($R_{\boldsymbol{\hat{y}}}$, $R_{\boldsymbol{\hat{z}}}$) as the lower bound of length of the bitstream when actual entropy encoding, according to Shannon rate-distortion theory~\cite{cover2012infotheory}. 
\begin{equation}
    R_{\boldsymbol{\hat{y}}} = \mathbb{E}[-\log_2P_{\boldsymbol{\hat{y}}}(\boldsymbol{\hat{y}})]  \quad   R_{\boldsymbol{\hat{z}}} = \mathbb{E}[-\log_2P_{\boldsymbol{\hat{z}}}(\boldsymbol{\hat{z}})]
\end{equation}
Thus, their entropy is introduced as one of the minimization objectives in the optimization process. At test time, estimated probability $P_{\boldsymbol{\hat{y}}}$ and $P_{\boldsymbol{\hat{z}}}$ are fed into the engine of entropy coder to obtain the binary bitstreams.

\begin{figure*}
  \centering
   \includegraphics[width=\linewidth]{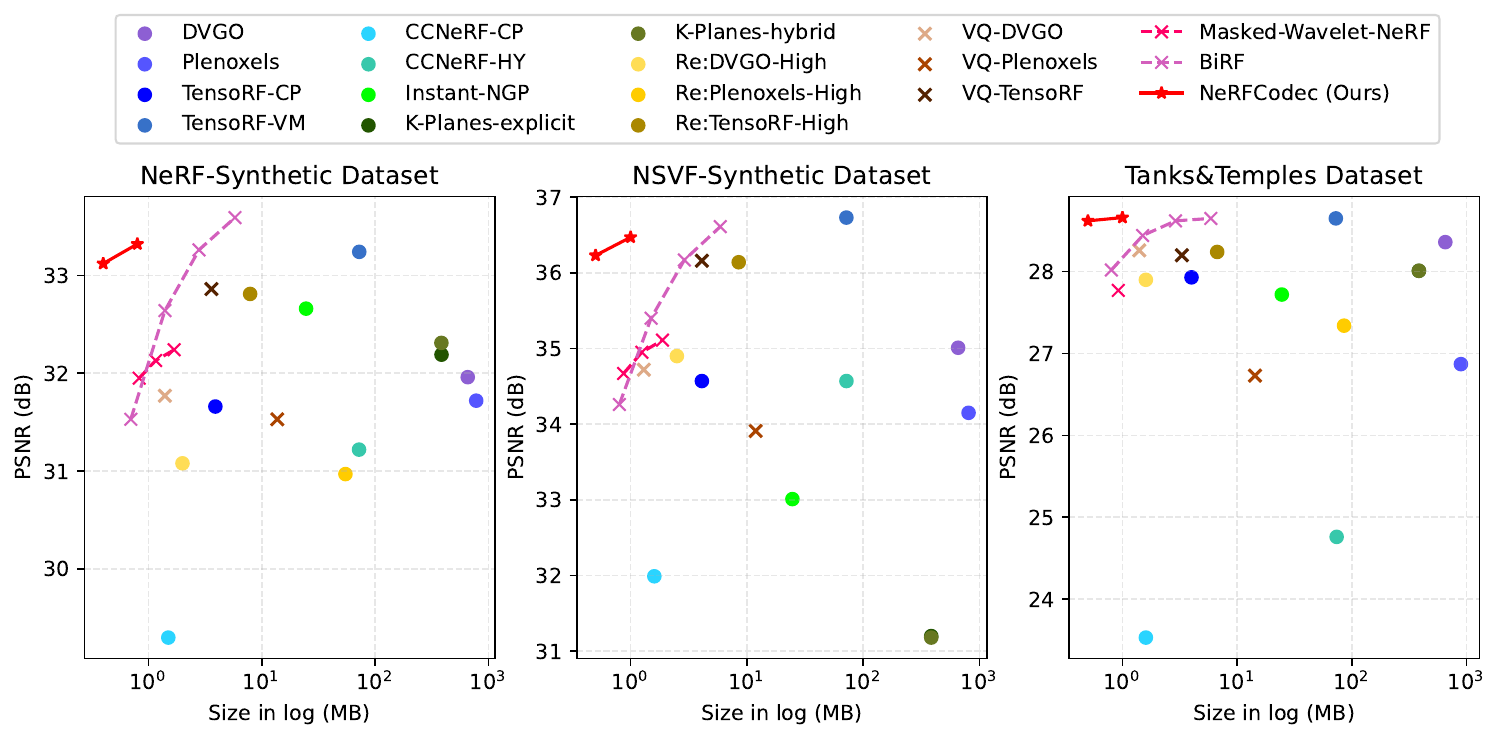}
   \vspace{-0.9cm}
   \caption{\textbf{Memory-quality plot}. We plot the memory footprint and visual quality score~(PSNR) to compare our method with baseline methods. We use different markers for our method and each category of baseline: circles ($\bullet$) for parameter-efficient data structure methods, crosses ($\times$) for parameter compression methods, and stars ($\star$) for ours.}
   \label{fig:rd_curve}
   \vspace{-0.6cm}
\end{figure*}

\subsection{Training}
\label{sec:training}
\boldparagraph{Loss Function}
In training, according to the volumetric rendering process described in \secref{sec:nerf}, we can render the pixel colors $\hat{C}(\mathbf{r})$  and supervise them with the ground truth RGB colors $C(\mathbf{r})$ along the sampled rays $\mathbf{r}$:
\begin{equation}
    \mathcal{L}_{\text{recon}} = \sum_{\mathbf{r} \in \mathcal{R}}||\hat{C}(\mathbf{r}) - C(\mathbf{r})||_2^2, \\
\end{equation}

The probability of the latent code is used to calculate its theoretical average minimum code length, which is the entropy of the latent code. 
The entropy of the latent code, serving as a proxy for the actual average encoding length, is incorporated into the final optimization objective to minimize the storage of latent code. 
\begin{equation}
    \mathcal{L}_{\text{entropy}} = R_{\hat{\by}} + R_{\hat{\bz}}
\end{equation}

The neural feature codec and small MLP for attribute regression will be jointly optimized by combining the rendering loss and the entropy loss.
The overall training loss for our system is defined as:
\begin{equation}
    \mathcal{L}_{\text{total}} = \mathcal{L}_{\text{recon}} + \lambda_{\text{entropy}} \mathcal{L}_{\text{entropy}}
\end{equation}

\boldparagraph{Multi-Stage Training}
First, we pre-train the TensoRF with 30,000 iterations to obtain feature planes~$\boldsymbol{x}$ and tiny MLP~$f$ for each scene.
Then, we perform a warm-up to the neural feature codec only with the feature plane reconstruction loss, leading to a better starting point for formal training.  
Next, we proceed with the joint training of the neural feature codec and the neural radiance field for 100,000 iterations with $\mathcal{L}_{\text{total}}$ as the loss function.
Finally, we perform quantization-aware training on the network parameters to be transmitted,\eg those from ``decoder head'' and ``MLP in renderer'', for 10,000 iterations.

\subsection{Test-Time Encoding and Decoding Process}
\label{sec:test-time}
At test time, the neural feature codec performs actual entropy coding and outputs bitstreams of latent code for transmission. In practice, the orthogonal feature planes are fed into the codec one by one to obtain their respective bitstreams of latent code. Apart from latent code, the content-specific network parameters in the neural feature decoder would be compressed and transmitted for the successful decoding of feature planes. Besides, other minority learnable parameters, such as MLPs, will be quantized and entropy-encoded into bitstreams. Thus, at the sending end, the bitstreams of different components, including latent code, content-specific parameters, small MLPs \etc, would be multiplexed into a total bitstream.
Then, at the receiving end, the total bitstream is first demultiplexed into multiple substreams representing different components. Each substream is independently entropy-decoded and reassembled to its corresponding location for reconstructing feature planes and volume rendering. 
It is worth noting that we only need to reconstruct feature planes once, and subsequent querying operations for each sampling point are executed on the reconstructed planes. Thus, the decoding operation will only incur a short decoding waiting time and will not affect the rendering efficiency. 

\subsection{Implementation Details}
\label{sec:detail}
Our method is compatible with different plane-based NeRF approaches. 
In practice, we choose two variants of plane-based NeRF, TensoRF-VM~\cite{Chen2022ECCV} and HexPlane~\cite{cao2023hexplane}, as the representative implementations of neural fields to validate compatibility. 
The implementation of neural feature codec is based on the CompressAI library~\cite{begaint2020compressai}. 
Specifically, we choose the neural image codec proposed by Cheng et al.~\cite{cheng2020learned} as the backbone of the neural feature codec for the following experiment session. 
In fact, other end-to-end image codecs~\cite{balle2018hyperprior, he2022elic} can also be substituted into our NeRFCodec framework.

\section{Experiments}
In this section, we evaluate our method's performance through quantitative comparisons with baselines, memory breakdown analysis, and thorough ablation studies to validate our design decisions. We mainly present the experimental results using TensoRF-VM as the neural field in this section and demonstrate the results using HexPlane as the neural field in the supplementary materials.

\boldparagraph{Datasets}
We evaluate our performance using both synthetic and real-world datasets. We use two synthetic datasets: the NeRF-Synthetic dataset~\cite{mildenhall2020nerf} and the NSVF-Synthetic dataset~\cite{LiuGLCT20nsvf}. We also choose the real-world dataset Tanks and Temples ~\cite{knapitsch2017tanks} and follow the NSVF evaluation protocol in scene selection and background masking.

\boldparagraph{Baselines}
We categorize baseline methods into two types: a)~methods focus on parameter-efficient data structure and b)~methods focus on parameter compression. Representatives of methods focusing on parameter-efficient data structure include DVGO~\cite{DVGO}, Plenoxels~\cite{yu2021plenoxels}, TensoRF~\cite{Chen2022ECCV}, CCNeRF~\cite{tang2022ccnerf}, Re:NeRF~\cite{deng2023renerf}, Instant-NGP~\cite{muller2022instant}, and K-Planes~\cite{fridovich2023kplanes}. Representatives of methods focusing on parameter compression include VQRF~\cite{li2023vqrf}, Masked Wavelet NeRF~\cite{rho2023maskedwavelet}, and BiRF~\cite{shin2023birf}.

\begin{figure}
    \small
    \setlength{\tabcolsep}{0pt}
    \def\mywidth{.25}
    \def\mysecwidth{.25}
    \begin{tabular}{cccc}
    \includegraphics[width=\mywidth\linewidth]{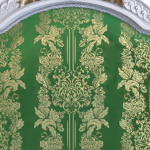} &
    \includegraphics[width=\mywidth\linewidth]{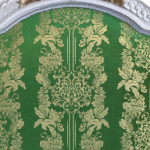} &
    \includegraphics[width=\mywidth\linewidth]{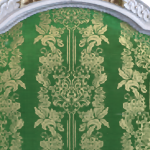} &
    \includegraphics[width=\mywidth\linewidth]{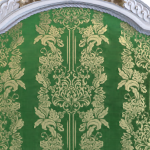} \\
    \includegraphics[width=\mywidth\linewidth]{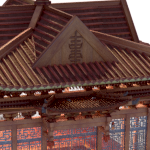} &
    \includegraphics[width=\mywidth\linewidth]{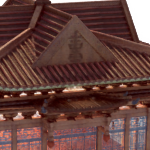} &
    \includegraphics[width=\mywidth\linewidth]{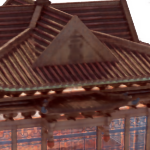} &
    \includegraphics[width=\mywidth\linewidth]{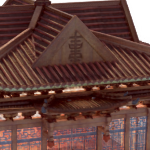} \\
    \includegraphics[width=\mysecwidth\linewidth]{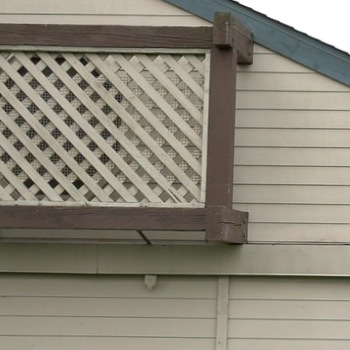} &
    \includegraphics[width=\mysecwidth\linewidth]{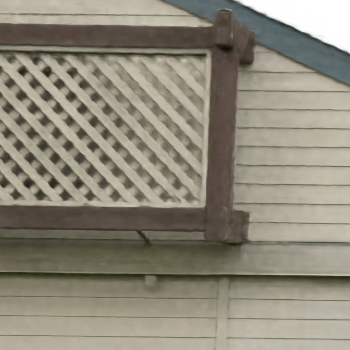} &
    \includegraphics[width=\mysecwidth\linewidth]{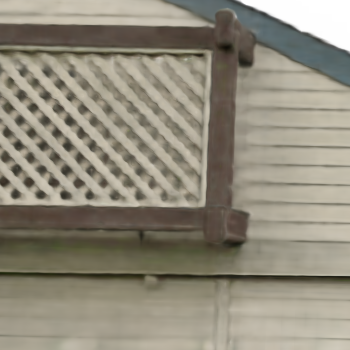} &
    \includegraphics[width=\mysecwidth\linewidth]{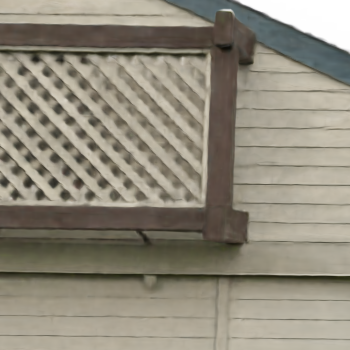} \\
    
    GT &
    Ours &
    \makecell{MWNeRF~\cite{rho2023maskedwavelet} } &
    TensoRF~\cite{Chen2022ECCV} \\
    \end{tabular}
    \vspace{-0.3cm}
    \caption{{\bf Qualitative comparison on NeRF-Synthetic.}  
    }
    \label{fig:qualitative}
    \vspace{-0.4cm}

\end{figure}

\subsection{Comparison with Baselines}
We report the average visual scores~(PSNR, SSIM) of test views and measure the average memory footprint of all scenes in each dataset. 
We plot the memory requirement and visual quality metrics of both our approach and the baseline method in \figref{fig:rd_curve}.
We leave the quantitative comparison of SSIM to the supplementary materials.
We achieve multi-bitrate points by using different numbers of channels in the feature planes.
In \figref{fig:rd_curve}, we observe that our method achieves superior rate-distortion performance compared to the baseline methods across three different datasets.
We observe an improvement compared to Masked Wavelet NeRF, which we attribute to the use of a non-linear transform instead of a wavelet transform. We also show the qualitative comparison in \figref{fig:qualitative}.

\subsection{Memory Breakdown}
In \tabref{tab:mem_breakdown}, we report the memory footprint of each component in the final bitstream, including the memory of compressed feature planes, compressed decoder adaptor, compressed MLP for attribute regression, and other side information.
We also report the original sizes of the uncompressed components and the corresponding compression ratios.
It's worth noting that the non-linear transform allows for a compression ratio over $1000$ times.
\begin{table}[!ht]
    \centering
    \small
    \vspace{-0.2cm}
    \begin{tabular}{p{2.5cm}ccc}
    \toprule
       & Compressed & Original & Ratio \\
       & (MB) & (MB) & \\
    \midrule
    Feature planes
    & 0.052  & 69.953  & $\times$1345.3  \\
    Decoder head
    & 0.269  & 1.761 & $\times$6.5  \\
    MLP in renderer
    & 0.042  & 0.160 & $\times$3.8  \\
    Feature vectors
    & 0.049  & 0.234 & $\times$4.8  \\
    Residual vectors
    & 0.041  & 0.234  & $\times$5.7  \\
    \midrule
    Total
    & 0.453  & 72.342  & $\times$159.7 \\
    \bottomrule
    \end{tabular}
    \vspace{-0.2cm}
    \caption{\textbf{Memory breakdown.}}
    \label{tab:mem_breakdown}
    \vspace{-0.3cm}
\end{table}

\subsection{Ablation Study}
We conduct ablation studies on NeRFCodec at a low rate point using \emph{Chair} scene from NeRF-Synthetic dataset. We leave more qualitative results to supplementary materials.

\qparagraph{Is content-adaptive encoder needed for compression?}
Our method follows an auto-encoder architecture, fine-tuning the feature encoder for each scene to obtain the latent code.
Alternatively, we explore a potential approach: bypassing the feature encoder and directly learning the latent code using the feature decoder, forming an auto-decoder architecture.
To compare these approaches, we implement the auto-decoder scheme by removing the feature encoder and setting the latent code as learnable parameters.
The remaining experimental settings, including quantization, entropy coding on the latent code, and decoder tuning strategy, remain consistent with the auto-encoder scheme.
For a fair comparison, we conduct a thorough hyperparameter search to identify the optimal learning rate for optimizing the latent code in the auto-decoder scheme.
In \tabref{tab:ablation_autodecoder}, we show the rate-distortion performance comparison between auto-encoder and auto-decoder schemes. 
The experimental results suggest that the auto-encoder scheme exhibits certain advantages in rate-distortion performance, possibly due to the feature encoder providing a more optimal initial value and optimization dynamics for the latent code.
\begin{table}[!ht]
    \centering
    \small
    \vspace{-0.1cm}
    \begin{tabular}{lccc}
    \toprule
    & Size~(MB) & PSNR~(dB) & SSIM \\
    \midrule
    auto-decoder& 0.461 & 34.41 & 0.973\\
    auto-encoder& 0.453 & 35.08 & 0.981\\
    \bottomrule
    \end{tabular}
    \vspace{-0.2cm}
    \caption{\textbf{Ablation on compression scheme.}}
    \label{tab:ablation_autodecoder}
    \vspace{-0.45cm}
\end{table}

\qparagraph{Can we re-use pre-trained 2D neural image codec without architecture modification?}
An alternative to leverage off-the-shelf neural image codec to compress plane-based hybrid NeRF is to split the feature planes into three channels, normalize, and feed it into the feature codec for compression. 
In this paradigm, the neural image codec could also be jointly fine-tuned with hybrid NeRF via rate-distortion loss. 
We report the experimental results in \tabref{tab:channel_wise_coding}. 
We observed that the bitrate consumption of this alternative is significantly higher than our approach. 
This could be attributed to the fact that they do not fully exploit inter-channel information during compression.
\begin{table}[!ht]
    \centering
    \small
    \vspace{-0.1cm}
    \begin{tabular}{lccc}
    \toprule
    & Size~(MB) & PSNR~(dB) & SSIM \\
    \midrule
    img. codec
    & 1.786 & 29.96 & 0.923\\
    tuned img. codec 
    & 1.150 & 33.25 & 0.961\\
    Ours
    & 0.453 & 35.08 & 0.981\\
    \bottomrule
    \end{tabular}
    \vspace{-0.2cm}
    \caption{\textbf{Ablation on re-using neural image codec.}}
    \label{tab:channel_wise_coding}
    \vspace{-0.45cm}
\end{table}

\qparagraph{Does the entropy loss contribute to storage saving?}
During training, we introduce entropy loss to constrain the predicted probability distribution by the probability model to be as close as possible to the actually unknown marginal distribution of the latent code.
When this entropy loss is minimized as much as possible during the optimization process, the actual code length during practical encoding will be close to the theoretically shortest code length.
We report the impact of entropy loss on the final storage savings of latent code in \tabref{tab:entropy_loss}. 
Compared to the scheme without entropy loss, we observed that the scheme with entropy loss significantly reduces the final code length. 
This highlights the importance of introducing entropy loss during training. 
\begin{table}[!ht]
    \centering
    \small
    \vspace{-0.1cm}
    \begin{tabular}{lccc}
    \toprule
    & Size~(MB) & PSNR~(dB) & SSIM \\
    \midrule
    w/o entropy loss & 0.701 & 34.98 & 0.980\\
    w/ entropy loss & 0.453 & 35.08 & 0.981\\
    \bottomrule
    \end{tabular}
    \vspace{-0.2cm}
    \caption{\textbf{Ablation on entropy loss.}}
    \label{tab:entropy_loss}
    \vspace{-0.45cm}
\end{table}

\qparagraph{Do we need a reconstruction loss for the feature plane?}
Intuitively, one might think that the reconstruction loss between the reconstructed feature plane and the pre-trained feature plane could benefit higher-quality NeRF reconstructions.
However, in practice, we find that adding a reconstruction loss to the reconstructed feature planes leads to degradation in rendering quality, reported in \tabref{tab:feat_rec_loss}. 
We hypothesize that the neural feature decoder can learn to synthesize features more suited for the following attribute regression without direct feature reconstruction loss.
\begin{table}[!ht]
    \centering
    \small
    \vspace{-0.1cm}
    \begin{tabular}{lccc}
    \toprule
    & Size~(MB) & PSNR~(dB) & SSIM \\
    \midrule
    w/ feature rec. loss & 0.479 & 32.81 & 0.965 \\
    w/o feature rec. loss & 0.453 & 35.08 & 0.981 \\
    \bottomrule
    \end{tabular}
    \vspace{-0.2cm}
    \caption{\textbf{Ablation on feature reconstruction loss.}}
    \label{tab:feat_rec_loss}
    \vspace{-0.45cm}
\end{table}

\qparagraph{Can a neural feature codec trained on a small-scale feature dataset generalize to new scenes?}
In cases where training the neural feature codec on a massive amount of feature planes is infeasible, we try to train a neural feature codec with a few scenes and test its generalization on unseen objects.
Specifically, we collect feature planes trained on eight objects from the NeRF-Synthetic dataset.
Next, we use these planes to train a neural feature codec on these scenes via rendering loss and entropy loss.
When evaluating on the training scenes, we find it could render reasonable images but with a slight degradation in visual scores.
However, when the codec trained on NeRF-Synthetic datasets is applied to the feature planes pre-trained from NSVF-Synthetic \textit{without any test-time optimization}, it fails to synthesize reasonable rendering results. 
The neural feature codec struggles to generalize with limited training data. Hence, our strategy of adapting well-trained 2D image codecs for NeRF compression holds value, considering the difficulty in obtaining large-scale plane feature datasets.

\section{Conclusion}
In this paper we propose NeRFCodec, an end-to-end compression framework for plane-based hybrid NeRF. 
The main idea is to leverage non-linear transform, quantization, and entropy coding for compressing feature planes in hybrid NeRF to achieve memory-efficient scene representation.
The experiments show that our method only uses a memory budget of 0.5 MB to represent a single scene while achieving high-quality novel view synthesis.
As a limitation, training the non-linear transform is time-consuming. 
Moreover, we need to train a specialized neural feature codec for each scene individually.
In the future, we plan to scale up the data collections of feature planes and train a generalized neural feature codec that generalizes well on unseen objects via training on large-scale datasets.

\small{\boldparagraph{Acknowledgements} This work is supported by the National Natural Science Foundation of China under Grant No.~U21B2004, No.~62071427, No.~62202418, Zhejiang University Education Foundation Qizhen Scholar Foundation, and the Fundamental Research Funds for the Central Universities under Grant No.~226-2022-00145. Yiyi Liao and Lu Yu are with Zhejiang Provincial Key Laboratory of Information Processing, Communication and Networking (IPCAN), Hangzhou 310007, China.}

\newpage
{\small
\bibliographystyle{ieee_fullname}
\bibliography{bibliography, bibliography_long, bibliography_custom}
}

\end{document}